\documentclass{sig-alternate}

\usepackage{subfigure}

\usepackage{algorithm}
\usepackage{algorithmic}

\usepackage{url}
\usepackage{verbatim}

\begin{document}

\title{User Clustering in Online Advertising via Topic Models}

\numberofauthors{3} 
\author{
\alignauthor
Sahin Cem Geyik \\
       \affaddr{Applied Science Division}\\
       \affaddr{Turn Inc.}\\
       \affaddr{Redwood City, CA, USA}\\
       \email{sgeyik@turn.com}
\alignauthor
Ali Dasdan\\
       \affaddr{Applied Science Division}\\
       \affaddr{Turn Inc.}\\
       \affaddr{Redwood City, CA, USA}\\
       \email{adasdan@turn.com}
\alignauthor
Kuang-Chih Lee *\\
       \affaddr{Computational Advertising}\\
       \affaddr{Yahoo! Labs}\\
       \affaddr{Sunnyvale, CA, USA}\\
       \email{leekc307@gmail.com}
}
\date{21 January 2014}

\newcommand{\theHalgorithm}{\arabic{algorithm}}

\maketitle
\begin{abstract}
In the domain of online advertising\let\thefootnote\relax\footnote{* This work was completed while the author was with Turn Inc.}, our aim is to serve the best ad to a user who visits a certain webpage, to maximize the chance of a desired action to be performed by this user after seeing the ad. While it is possible to generate a different prediction model for each user to tell if he/she will act on a given ad, the prediction result typically will be quite unreliable with huge variance, since the desired actions are extremely sparse, and the set of users is huge (hundreds of millions) and extremely volatile, i.e., a lot of new users are introduced everyday, or are no longer valid. In this paper we aim to improve the accuracy in finding users who will perform the desired action, by assigning each user to a cluster, where the number of clusters is much smaller than the number of users (in the order of hundreds). Each user will fall into the same cluster with another user if their event history are similar. For this purpose, we modify the probabilistic latent semantic analysis (pLSA) model by assuming the independence of the user and the cluster id, given the history of events. This assumption helps us to identify a cluster of a new user without re-clustering all the users. We present the details of the algorithm we employed as well as the distributed implementation on Hadoop, and some initial results on the clusters that were generated by the algorithm.
\end{abstract}
\category{J.0}{Computer Applications}{General}
\\ \category{I.5.3}{Computing Methodologies}{Pattern Recognition}[Clustering]

\terms{Algorithms, Application}

\keywords{Online advertising, User clustering, Topic models}

\section{Introduction}	\label{sec:intro}
In online advertising, the goal is to find the best ad under constraints for a given user in an online context. The context varies based on the applicable ad format (e.g., banner or display ad, video ad) and the device (e.g., desktop, mobile) the user is using at the time of the ad impression (showing the ad to user). The constraints are mainly imposed by advertisers on the target user and context parameters, e.g., users of certain age range visiting webpages of certain categories. The expectation of an advertiser is one of: brand recognition, clicks, or actions. Actions are advertiser defined and can be one of inquiring about or purchasing a product, filling out a form, visiting a certain page, etc.~\cite{klee_2012}.

A certain ad from an advertiser can be shown to a user in an online context only if the value for the ad impression opportunity is high enough to win in a real-time, competitive auction~\cite{borgs_2007}. Advertisers directly or indirectly through demand-side platforms, entities that work on behalf of advertisers to deal with real-time bidding ad exchanges, signal their value via bids. The bid for an ad impression is calculated as the action probability given a user in a certain online context multiplied by the cost-per-action goal an advertiser wants to meet or beat.

The action probability computation is about computing the likelihood of an action by a certain user on a certain ad on a certain webpage. Since it is impossible for every user to see every ad on every webpage, the likelihood is almost always unknown. As a result, we need to use campaign/website/user hierarchies and other techniques to reduce this extreme sparsity and compute this extremely rare event~\cite{klee_2012}. The idea can be explained best by an example: The likelihood of a user visiting a certain webpage can be unknown or zero but we may have better confidence if we consider the likelihood of the same user visiting the top level domain that the webpage belongs to (e.g. finance page of a news website vs. any page under the same news website). The example that applies to this paper is along the user dimension: The likelihood of a user acting on a certain ad on a certain webpage can be unknown but we may approximate the likelihood if we consider the actions of similar users on the same or similar webpages or the top level domains they belong to.

User similarity can be deduced from user clustering. User clustering in turn can be done utilizing user attributes such as age, gender, income, geographic attributes; however, most of the user attributes come from third-party data providers, who provide the data to advertisers at a cost (and only for a subset of users). As a result, we have to resort to using user attributes that do not come from these data providers.

\begin{figure}[htb]
\centering
\includegraphics[width=3.2in]{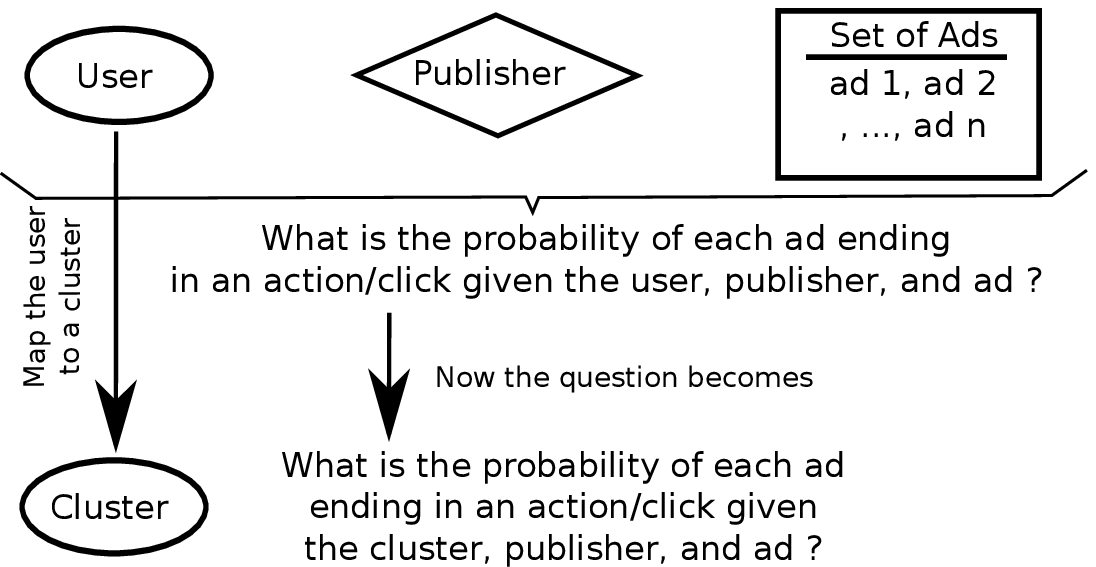}
\caption{Problem Description}
\label{fig:figure_introduction_problem}
\end{figure}

In this paper, we propose to use the event history of users to cluster users into multiple classes and use the class id as a feature combined with other features, such as advertiser and publisher properties, to calculate the action probability at the time of ad serving. This clustering process, named \emph{user segmentation} in the advertising domain, is receiving increasing attention. We present current efforts in literature in \S~\ref{sec:previous_work}. Please see Figure~\ref{fig:figure_introduction_problem} for a description of the problem we are attempting to solve. The event history is built by the event tracking process. When a user performs the designed action in the advertiser conversion page, a small piece of java script code, called \emph{beacon} or \emph{tracking pixel}, fires an event to the ad serving system to report that a desired action has been fulfilled. Since we use the beacon id to identify each event in our implementation, we will refer to events as beacons for the rest of the paper.

The methodology we propose in this paper is inspired highly by  \emph{probabilistic latent semantic analysis} (pLSA)~\cite{hofmann_2004}, which has been used in document clustering, scene classification~\cite{bosch_2006}, and user clustering for news recommendation~\cite{das_2007}. However, there is a significant difference between pLSA and our approach. While in pLSA the user and the item (e.g., recommended item for recommender systems, or generated words for document clustering) are independent of each other given the cluster information, in our case, the cluster id and the user id are independent of each other given a previous event of a user. Such a difference actually helps to determine the cluster which a new user belongs to at run-time just by looking at its beacon history. On the contrary, pLSA requires a huge amount of memory to store the cluster id for each existing user, and has to recompute the topic model from scratch for new users. Although the modification proposed for pLSA is pretty general and can be applied to multiple domains, we believe it especially is useful for online advertisement domain (due to the dynamic nature, i.e. constantly evolving set of users). This modification also significantly reduces the computation of the expectation maximization (EM) algorithm that is used to learn the clustering parameters.

The rest of the paper is as follows. We first give the relevant previous work for both user clustering, as well as its different applications, in \S~\ref{sec:previous_work}. Then, we introduce our methodology, as well as providing the differences compared to pLSA and our motivation in these differences in \S~\ref{sec:methodology}. We give the implementation details in \S~\ref{sec:implementation}. The initial results of our clustering methodology is provided in \S~\ref{sec:results}. Finally, we conclude the paper and propose future work in \S~\ref{sec:conclusion}.

\section{Previous Work} \label{sec:previous_work}

User clustering (or user segmentation) for better performance in advertising systems is a subject that has gained increasing attention in the past years. An analysis of how user segmentation can help click-through rate (CTR), i.e. the percentage of the advertisements that are clicked by a user, has been presented in \cite{yan_2009}, which justifies this interest.

One of the most common methodologies for grouping similarly behaving users in the online advertising domain is the utilization of \emph{topic models}. \cite{wu_2009} introduces Probabilistic Latent Semantic User Segmentation (PLSUS) for advertising, where each user is represented as a bag of words according to his/her search queries. Their methodology applies pLSA, hence suffers from the same problems aforementioned (new users and cost of computation). Latent Dirichlet Allocation (LDA) is also another type of topic model that is used commonly to cluster users for many applications. \cite{fujimoto_2011} gives an example where web users are clustered, assuming the users are the documents and the visited websites are the words. Furthermore, \cite{tu_2010} gives an application of LDA for online advertising (but on a small, sampled set of users). The main problem with LDA is the computational complexity to train them, and the difficulty in assignment of a new user to a cluster (topic). While works such as \cite{newman_2009} present distributed implementations of Latent Dirichlet Allocation and the Hierarchical Dirichet Process as fast ways to train topic models, implementation overhead (cost of human resources for implementation, as well as scalability issues that affect applicability on a real advertising scenario) is not negligible.

Other than directly segmenting the users, there has also been work on estimating clicks or actions via the user properties. The first case we would like to list, similar to our work, is given in \cite{aly_2012}, where the authors utilize user actions to classify users into two groups for each advertiser: \emph{converters} (will purchase, or click etc.) and \emph{non-converters}. They utilize support vector machines (SVM), and train this model using the past actions of users (similar to the \emph{beacons} we mention in this paper). The problem with this approach is that it effectively classifies users into two groups, and such grouping may not be valid for all campaigns (i.e. converter for an ad for a certain product may not be a converter for another product). Our clustering approach aims at separating users to multiple groups, and calculating action/click probabilities for each group, over many campaigns.

Another approach in utilizing user information for advertising is given in \cite{liu_2011}, where the authors employ friendship data (due to similarity of friends, i.e. \emph{homophily}) as well as other information that arises in a large online social networking site. This kind of information however is often proprietary, and not available to most online advertisers. The paper in \cite{joshi_2011} aims at incorporating user demographic information into a text representation. This representation is later utilized for matching relevant ads to both ad and publisher's textual properties to enhance advertising efficiency. While the authors show a CTR lift due to their methodology, it is not certain whether such matching is feasible or useful in online advertising systems where there is only limited demographic information for users as well as publishers (textual features such as title, keywords etc. are not always available, but are crucial for the methodology presented in the paper).

Finally, in mobile advertising domain, \cite{ma_2012} suggests the usage of mobile user patterns and therefore segmenting users based on the pattern similarity. The authors utilize categorization of location and activities as well as a constraint-based Bayesian Matrix Factorization model to deal with sparsity. Main problems with this approach are due to privacy and scalability, i.e. the lack of detailed activity information available to advertising systems, and the number of users in the advertising domain that make it extremely hard for such methods to be applied.

\section{Methodology} \label{sec:methodology}
As aforementioned, our clustering algorithm is highly inspired by the \emph{probabilistic Latent Semantic Analysis} (pLSA) \cite{hofmann_2004}, with a difference in the independence assumption of user and cluster, given the history of the user. In this section, we first give an introduction to pLSA, and then give the difference of our model and our motivation to why we chose that path. We conclude this section with the algorithmic and mathematical details of the model we utilized.

\subsection{Probabilistic Latent Semantic Analysis} 
Probabilistic latent semantic analysis~\cite{hofmann_2004} assumes the existence of a latent class variable (i.e. clusters, or topics), and that each document has a distribution over the set of topics. Furthermore, it assumes that given the topic information, probability of a word being generated by a document is independent of that document's id:
\begin{equation}
p(\textrm{word}_\textrm{k} | \textrm{document}_\textrm{j} , \textrm{topic}_\textrm{i}) = p(\textrm{word}_\textrm{k} | \textrm{topic}_\textrm{i}).
\end{equation}
This makes pLSA a really powerful generative model. That is, if we want to generate a new word from a document $d$, we need to first generate the topic, or cluster $c$ (according to p(c$|$d)), and then pick up a word $w$ (according to p(w$|$c)). Mapping of this to a recommendation system is by first selecting the cluster for a user (document) and then generating the item (word, i.e. recommendation) from this cluster.

The parameters of the pLSA model is learned using the expectation maximization algorithm~\cite{hofmann_2004} as follows:
\begin{itemize}
\item \textbf{Expectation Step:} Update the probability of a cluster given the document and word:
\[
p(c_i|d_j,w_k) = \frac{p(c_i,d_j,w_k)}{\sum_{c_m} p(c_m,d_j,w_k)} ~~~~~~~~~~~~~~~~~~~~~~~~~~~~~~~
\]
\[
~~ =  \frac{p(w_k|c_i) ~ p(c_i|d_j) ~ p(d_j)}{\sum_{c_m} p(w_k|c_m) ~ p(c_m|d_j) ~ p(d_j)}
\]
\[
 =  \frac{p(w_k|c_i) ~ p(c_i|d_j)}{\sum_{c_m} p(w_k|c_m) ~ p(c_m|d_j)} ~ . ~~~
\]
\item \textbf{Maximization Step:} Update the conditional probabilities of words given clusters, and clusters given documents:
\[
p(w_k|c_i) = \frac{\sum_{d_j} p(c_i,d_j,w_k)}{\sum_{w_m} \sum_{d_j} p(c_i,d_j,w_m)} ~~~~~~~~~~~~~~~~~~~~
\]
\[
~~~~~ = \frac{\sum_{d_j} \alpha p(d_j,w_k) ~ p(c_i|d_j,w_k)}{\sum_{w_m} \sum_{d_j} \alpha p(d_j,w_m) ~ p(c_i|d_j,w_m)}
\]
\[
~~~ = \frac{\sum_{d_j} n(d_j,w_k) ~ p(c_i|d_j,w_k)}{\sum_{w_m}\sum_{d_j} n(d_j,w_m) ~ p(c_i|d_j,w_m)}
\]
where $n(d_j,w_k)$ is the number of times $d_j$ and $w_k$ has been seen together. This number arises when $\alpha$ is taken to be the total number of word-document co-occurences (i.e. $p(d_j,w_k) = \frac{n(d_j,w_k)}{\sum_{d_n}\sum_{w_m}n(d_n,w_m)}$, hence $\alpha = \sum_{d_n}\sum_{w_m}n(d_n,w_m)$).

~~

Furthermore,
\[
p(c_i|d_j) = \frac{\sum_{w_k} p(c_i,d_j,w_k)}{\sum_{c_m} \sum_{w_k} p(c_m,d_j,w_k)} ~~~~~~~~~~~~~~~~~~~~~~~~~~~~~~~~
\]
\[
= \frac{\sum_{w_k} p(d_j,w_k) ~ p(c_i|d_j,w_k)}{p(d_j)} ~~~~~~~~~~~
\]
\[
= \frac{\sum_{w_k} \alpha p(d_j,w_k) ~ p(c_i|d_j,w_k)}{\alpha p(d_j)} ~~~~~~~~~
\]
\[
= \frac{\sum_{w_k} n(d_j,w_k) ~ p(c_i | d_j, w_k)}{n(d_j)} ~~~~~~~~~~
\]
where $n(d_j)$ is the number of times we have seen a word from document $d_j$ in the training data. This number arises when $\alpha$ is again taken to be the total number of word-document co-occurences (i.e. $p(d_j) = \frac{\sum_{w_m}n(d_j,w_m)}{\sum_{d_n}\sum_{w_m} n(d_n,w_m)}$, hence $\alpha = \sum_{d_n}\sum_{w_m} n(d_n,w_m)$ and $n(d_j) = \sum_{w_m}n(d_j,w_m)$. Please note that the $\alpha$ value also causes $p(d_j,w_k)$ above the fraction to be $n(d_j,w_k)$.
\end{itemize}
As aforementioned, the pLSA model is a very appropriate model for recommendation systems, since it is capable of suggesting a new item (i.e. one that the user has not seen before) to a user, due to the assumption that given the cluster id, the user and the item sets are independent of each other. But it comes with a cost, first being the necessity to store the cluster information for each user, and second the problem of determining cluster for the new users. In most recommendation systems, the user sets are quite stable (i.e. not too many users come and go), hence it is usually unlikely we will have to determine a user's cluster after the training period is over. Even if the item lists change frequently, it is a much easier task to update the item probabilities for each cluster (a single maximization step), as long as the user sets do not change too frequently, in which case we need to run the whole EM-step.

We will next give the difficulties that arise in the domain of online advertising, hence the need for a different approach.

\subsection{Challenges of Online Advertising}
As we have explained in the previous subsection, pLSA works well in recommending new items, given the users are stable and we know the cluster id for these users. In online advertising however, we try to provide to all the users over all the web (hence no subscription system), and the user ids are much more volatile, since a web user's id is based on the cookies on his/her device. If these cookies are deleted, no longer is his cluster id valid, since the user id no longer exists as well. Also note that this causes many new users arrive in (or leave) the ecosystem every day. What we need is a system that looks at the history of the user's events and assign it to a cluster at the ad-serving time (i.e. run-time).

Another problem in online advertising domain is the fact that we are not trying to recommend new items to users, at least not in the context of this paper. What we are trying to do is to be able to understand the action/click probability of a user, given that it belongs to a certain cluster (i.e. if a user belongs to a cluster due to his/her events, we want its cluster id, not what other events (beacons) it will perform, i.e. a generalization given its past events (beacons)). Hence we want a model which \emph{only} looks at the event history of a user (not its user id, since it is too volatile) and give the group (cluster) id (this also covers all the new users, as long as we know their beacon history), so that we can calculate the action probability of that group (cluster). This also gives us the biggest difference of our methodology compared to pLSA: Given the beacon history, the user is independent of the cluster id. This difference, as well as others that arise from this, are given in the next subsection.

\subsection{Differences of Our Methodology\\Compared to pLSA}
The biggest difference of the clustering method we employ compared to pLSA is in the model. Keeping with the terms of online advertising, while pLSA assumes the independency of the \emph{beacons} given the user id, as long as the cluster id is known, we assert that the cluster id is independent of user id, given the beacon id. This independence assumption in the model helps us in determining the cluster of a new or existing user at run-time due to its beacon history, therefore, we do not need to store any cluster id for any user. Since now we can determine the cluster of a user at run-time, an ad-serving system tries to calculate the action or click probability of this user given a specific ad on a specific web-site (publisher) as ``\emph{What is the action/click probability of a user belonging to cluster i for this ad at this website?}''.

\begin{figure}[!t]
\centering
\subfigure[Plate notation for pLSA]{ \includegraphics[width=2in]{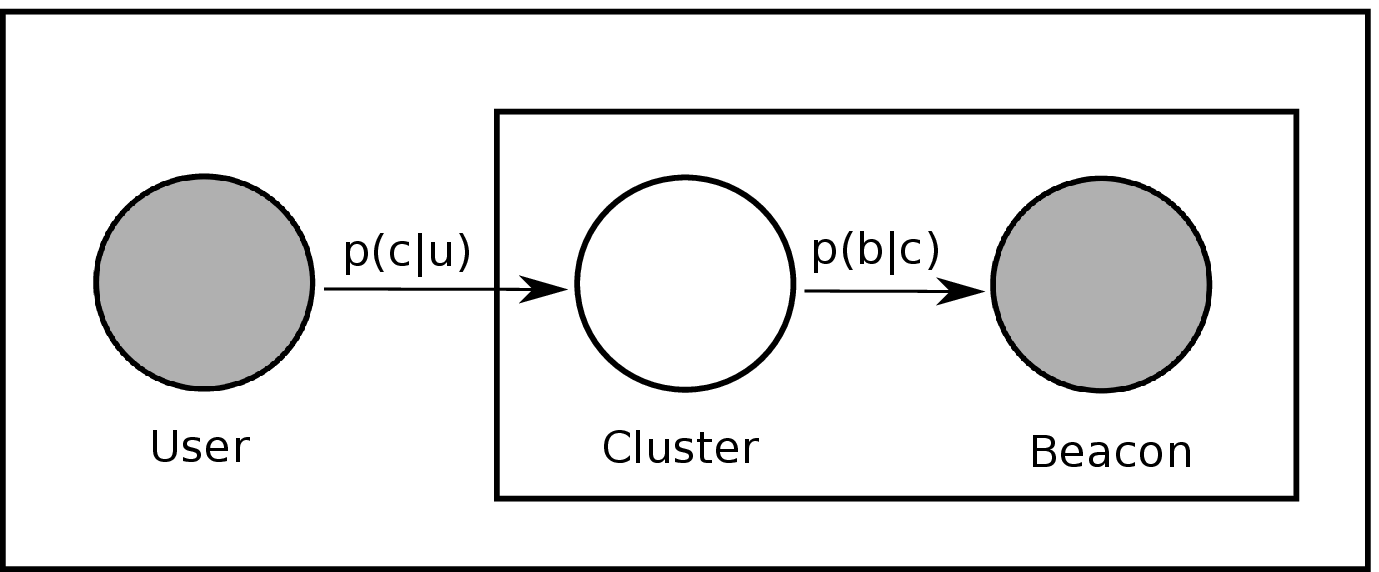} \label{subfig:pLSA_plate}} \\
\subfigure[Plate notation for the clustering method we utilized]{ \includegraphics[width=2in]{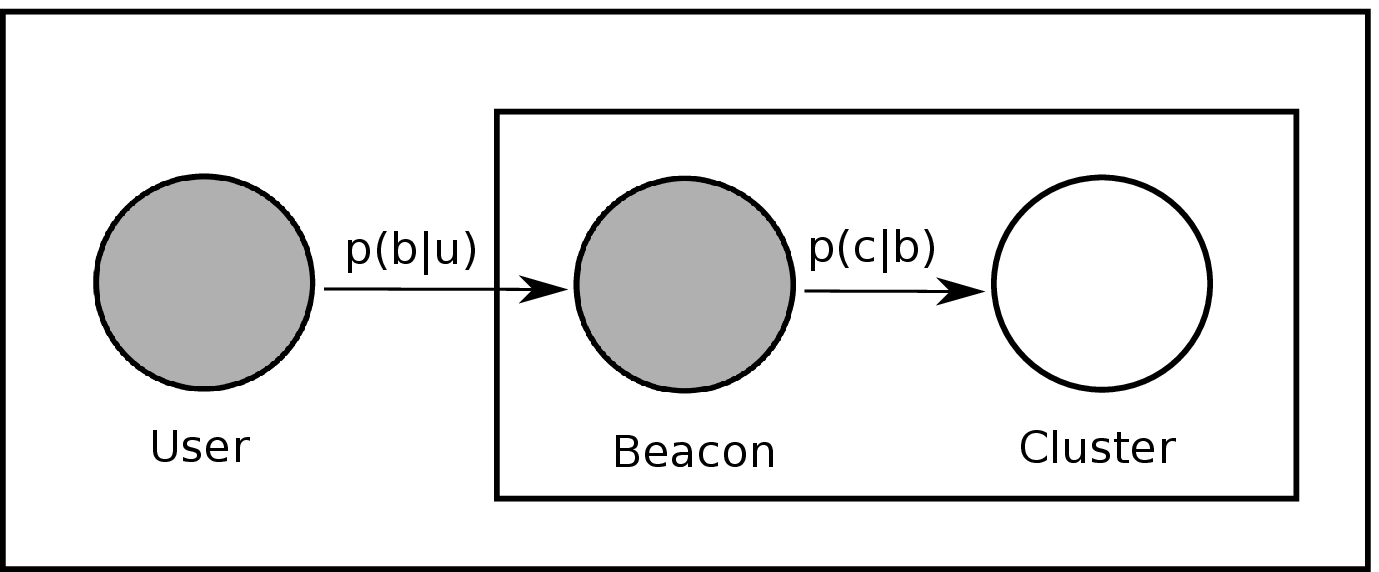} \label{subfig:our_plate}}
\caption{Difference between pLSA and the clustering we utilized. Please notice that while pLSA assumes that the beacons and users are independent given the cluster id, we assume that the users and clusters are independent given a beacon id from the user's history of events.}
\label{fig:difference_vs_plsa}
\end{figure}

Please see the difference of the models as a plate notation in Figure~\ref{fig:difference_vs_plsa}. In the figure, it can be seen that the user notation could be directly removed from our model, and hence its relationship with pLSA becomes pretty weak (for good reason too, due to the nature of online advertising as we listed in previous section). Furthermore, our model is pretty general and can be applied to different application domains, however it is especially necessary and useful in online advertising due to constantly evolving set of users.

Our proposed formulation naturally also brings some differences in the EM algorithm that needs to be run for training model probabilities. As an obvious example of this, we had mentioned previously that the expectation step of pLSA was calculating $p(c_i|d_j,w_k)$ which is the conditional probability of a cluster i given document j and word k, which can be interpreted into the online advertising domain as conditional probability of cluster i given user j ($u_j$) and beacon k ($b_k$). But due to our model's independence assumption, $p(c_i|u_j,b_k) = p(c_i|b_k)$. The details of the EM algorithm we employ, as well as the determination of cluster for a user at run-time is given in the next subsection.

\subsection{Training the Model and Run-time Cluster Determination for Users}
To be able to determine the cluster of a user at run-time, we need to have access to the beacon history of this user. The beacon history of a user is the set of beacons (events) that have been performed by this user in the previous x days (we took this value to be 60 days for our implementation), e.g. $h(u_i) = \{b_{i,1},b_{i,2},...,b_{i,n}\}$ for user i, if it has performed n actions/beacons. Not all of these n beacons have to be unique, hence we can come up with a probability distribution such as $p(b_j|u_i) = \frac{n(b_j,u_i)}{n(u_i)}$, where $n(b_j,u_i)$ is the number of times user i has seen beacon j, and $n(u_i)$ is the total number of beacons that user i has seen. Therefore, in run-time, we calculate below:
\begin{equation}
\label{eq:ciuj}
p(c_i|u_j) = \sum_{b_k} p(c_i,b_k|u_j) = \sum_{b_k} p(c_i|b_k) p(b_k|u_j)
\end{equation}
for each cluster i, and determine the cluster that gives the maximum conditional probability to be the cluster that user j belongs to. Since the beacon probabilities given users are known during run-time, all we have to store is the beacon to cluster mapping (i.e. $p(c_i|b_k)$ for all beacon cluster pairs). Please note that this requires much less storage than keeping cluster ids for each user ($\sim$500 million users, while we take $\sim$600 clusters, due to our previous experience with different numbers of clusters, and around 13,000 beacons, and not every beacon has above zero conditional probability for each cluster). It is also robust to changing beacon history for each user, as well as new users (whose beacons might have been fired elsewhere, or these users were not present when model training was being performed).
\\The EM algorithm is trying to maximize
\\ $\prod_{\forall u_j} \prod_{\forall c_i\textrm{ where }p(c_i|u_j)>0} p(c_i|u_j)$
\\= $\prod_{\forall u_j} \prod_{\forall c_i\textrm{ where }p(c_i|u_j)>0} \sum_{b_k} p(c_i|b_k) p(b_k|u_j)$, per Eq.~\ref{eq:ciuj}.
\\This process is directed by the above cluster determination policy, and is as follows:
\begin{itemize}
\item \textbf{Expectation Step:} For each user in the training data, calculate the probability of this user belonging to a cluster i:
\[
p(c_i|u_j) = \sum_{b_k} p(c_i|b_k) p(b_k|u_j), ~~~~~~~~~~~~~~~~~~~~~~~~~~~~~~~~~~
\]
which can further be written as:
\[
p(c_i|u_j) = \sum_{b_k} \frac{p(c_i) ~ p(b_k|c_i)}{p(b_k)} p(b_k|u_j), ~~~~~~~~~~~~~~~~~~~~~~~~~~~
\]
where $p(b_k)$ is the probability of beacon k, and is calculated as $p(b_k) = \frac{\sum_{u_n} n(u_n,b_k)}{\sum_{u_n} \sum_{b_m} n(u_n,b_m)}$, where $n(u_n,b_k)$ is the number of times the beacon k has been seen by user n in the training data.
\item \textbf{Maximization Step:} In the maximization step, we calculate the values $p(c_i)$ and $p(b_k|c_i)$:
\[
p(c_i) = \sum_{u_j} p(u_j) ~ p(c_i|u_j), ~~~~~~~~~~~~~~~~~~~~~~~~~~~~~~~~~~~~~~~~~~~~~~
\]
where $p(c_i|u_j)$ is the output of the expectation step. Furthermore, the marginal probability for the user u$_i$ is $p(u_i)=\frac{n(u_i)}{\sum_{\forall u_j} n(u_j)}$ where $n(u_i)$ is the number of beacons $u_i$ has fired. We also calculate:
\[
p(b_k|c_i) = \sum_{u_j} p(b_k|u_j) p(u_j|c_i) ~~~~~~~~~~~~~~~~~~~~~~~~~~~~~~~~
\]
\[
= \sum_{u_j} p(b_k|u_j) \frac{p(c_i|u_j) ~ p(u_j)}{p(c_i)} ~~~~~~~~~~~~
\]
where $p(c_i|u_j)$ is the output of the expectation step, and $p(c_i)$ is calculated as part of the maximization step.
\end{itemize}
The above formulation means that in the expectation step, we assign users to clusters (according to beacon cluster mapping, and since we do not have this at the beginning, randomly); and in the maximization step we calculate beacon to cluster mapping parameters, as well as the cluster probabilities. This goes on until convergence (i.e. the cluster probabilities as well as beacon to cluster probabilities no longer change significantly). This scheme is detailed in Algorithm~\ref{alg:em_for_ours}. One detail in the algorithm is the difference between hard clustering and soft clustering. In hard clustering, we assign the user to the most likely cluster $c_i$, making $p(c_i|u) = 1$ (similar to run-time cluster determination); while in soft clustering we assign the user to multiple clusters where $\sum_{c_i} p(c_i|u) = 1$. 

\begin{algorithm} [t!]
\caption{EM Algorithm to Learn the Parameters of the User Clusters}
\label{alg:em_for_ours}
\begin{algorithmic}
\STATE{\textbf{learnClusterParameters}}
\STATE{~~~~~~~~~~ (userSet $U$, beaconSet B, numClusters):}
\FOR{each user $u_j \in U$}
  \STATE{assign $u_j$ to a random $c_i$ where $i \in [1,\textrm{numClusters}]$}
  \STATE{$p(c_i|u_j)=1$}
\ENDFOR
\FOR{each cluster $c_i$}
  \STATE{$p_{old}(c_i)$ = null}
  \STATE{calculate $p_{new}(c_i)$}
  \FOR{each $b_k$}
    \STATE{$p_{old}(b_k|c_i)$ = null}
    \STATE{calculate $p_{new}(b_k|c_i)$}
  \ENDFOR
\ENDFOR
\WHILE{\\\{$p_{old}(c_i)$ and $p_{new}(c_i)$ are significantly\\ ~~~~~ different for any $c_i$\} \\ \textbf{~~~~~~~~ OR} \\ \{$p_{old}(b_k|c_i)$ and $p_{new}(b_k|c_i)$ are significantly\\ ~~~~~ different for any $c_i$ and $b_k$\}}
  \FOR{each user $u_j$}
    \IF{Soft Clustering}
      \STATE{Calculate $p(c_i|u_j)$ for all $c_i$}
    \ELSE
      \STATE{Choose $c_i$ for $u_j$ where $p(c_i|u_j)$ is maximum}
      \STATE{$p(c_i|u_j)=1$}
    \ENDIF
  \ENDFOR
  \FOR{each cluster $c_i$}
    \STATE{$p_{old}(c_i)$ = $p_{new}(c_i)$}
    \STATE{calculate $p_{new}(c_i)$}
    \FOR{each $b_k$}
     \STATE{$p_{old}(b_k|c_i)$ = $p_{new}(b_k|c_i)$}
     \STATE{calculate $p_{new}(b_k|c_i)$}
    \ENDFOR
  \ENDFOR
\ENDWHILE
\FOR{each cluster $c_i$}
    \FOR{each beacon $b_k$}
      \STATE{Calculate and return $p(c_i|b_k)$}
    \ENDFOR
\ENDFOR
\end{algorithmic}
\end{algorithm}
At the end of EM process, we output $p(c_i|b_k)$ for each beacon cluster pair, to be used in run-time user to cluster determination.

\section{Implementation Details} \label{sec:implementation}
\begin{figure}[!t]
\subfigure[Code for User-Cluster Determination]{ \includegraphics[width=2.4in]{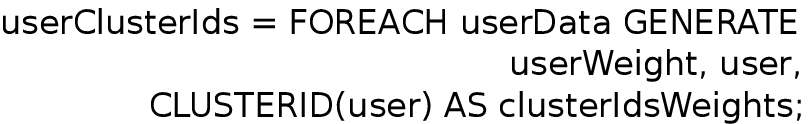} \label{subfig:figure_pig_code_1}} \\
\subfigure[Code for Probability Normalization]{ \includegraphics[width=2.5in]{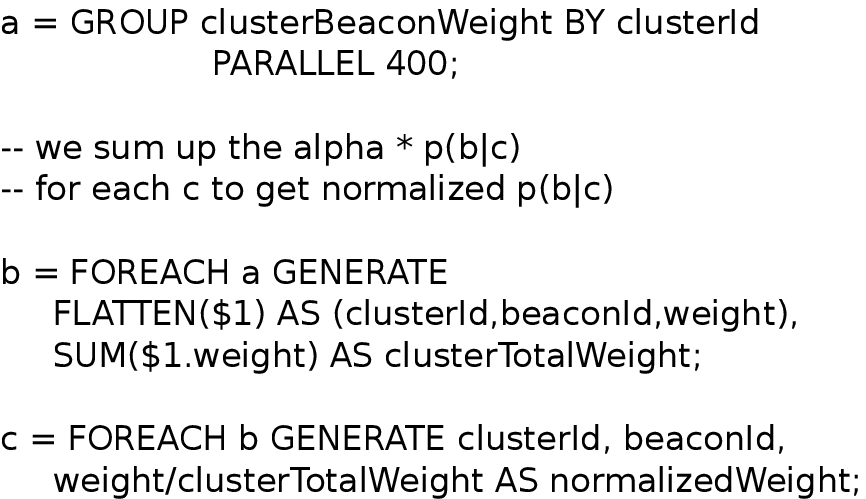} \label{subfig:figure_pig_code_2}}
\caption{Apache Pig code examples from our user clustering implementation, which should be helpful in understanding the parallel computation steps.}
\label{fig:pig_code_1}
\end{figure}

We utilized Apache Pig~\cite{pig_2008} to implement the clustering algorithm outlined in this paper. Pig is a high-level language working on top of Hadoop~\cite{hadoop_2012} framework, hence makes our clustering task complete much quicker due to parallelization. As stated in the previous section, the clustering task starts with determining cluster id for each line (i.e. data for a single user) in the user information folder, that contains $p(u_j)$ as well as all the set of unique beacons this user has encountered, with their conditional probabilities ($p(b_k|u_j)$). This user data is pre-calculated to compress the whole training data, as well as to reduce the time for a single EM step. The pig script section for this cluster determination step is given in Figure~\ref{subfig:figure_pig_code_1}. In the figure, \emph{userWeight} is $p(u_j)$, user is a bag of $p(b_k|u_j)$, \emph{CLUSTERID} is a user defined function (udf) which has the inputs of $p(c_i)$ and $p(b_k|c_i)$ as well as the \emph{user} (bag of $p(b_k|u_j)$) and produces a bag of $p(c_i|u_j)$ (i.e. \emph{clusterIdsWeights}). This structure means that we can calculate the EM step parameters by flattening (unfolding) either the \emph{user} bag, or \emph{clusterIdsWeights}.

Another problem is the calculation of normalized probabilities, where at the beginning we have a weight for each probability, but we need the normalized values for the same group. For example, suppose in the below example, we start with a variable \emph{clusterBeaconWeight} which has for each ($c_i$,$b_k$) pair the value $\alpha p(b_k|c_i)$ (where $\alpha$ is not known). To calculate the exact value $p(b_k|c_i)$, we need to first group according to cluster id, then sum up $p(b_k|c_i)$ as the normalizing factor. Later, we unfold the group with this sum appended, and then divide by the normalizer to find the exact probability. The example code piece that achieves this flow is given in Figure~\ref{subfig:figure_pig_code_2}, where the number after \emph{PARALLEL} gives the number of reducers, and the keyword \emph{FLATTEN} is used to unfold the bag which is the set of $p(b_k|c_i)$ for the same cluster.

The power of using Pig, hence the Hadoop framework is due to assigning the cluster ids to many users in different machines, as well as the sum operations working on different cluster parameters at the same time. Please note that the algorithm (Algorithm~\ref{alg:em_for_ours}) we provided for model training is \emph{embarrassingly parallel}, i.e. at each point of training we can divide the dataset completely according to current formulation we want to compute (whether by user id, beacon id, or cluster id). This also means that our parallel implementation is \emph{linearly scalable}, i.e. it is $n$-times faster than a single machine approach at any step of training$^*$\let\thefootnote\relax\footnote{* In Algorithm~\ref{alg:em_for_ours}, calculation of $p(c_i|u_j)$ is one map-reduce job (mapper-intensive), and calculation of both $p_{new}(c_i)$ and $p_{new}(b_k | c_i)$ is another separate map-reduce job (reducer-intensive). These two run sequentially and repeatedly until convergence.}, where $n$ is the number of machines (e.g. number of reducers for reducer-side operations such as grouping and counting, or number of mappers for mapper-side operations, such as calculating $p(c_i | u_j)$ in Algorithm~\ref{alg:em_for_ours}). It all adds up to be able to learn the model feasibly for the huge number of users and events we deal with in online advertising domain. In our model, we utilized a set of $\sim$13,000 beacons that have been captured over 3 months. This set has been selected by first removing the both tails from the whole set of beacons, i.e. removing the most frequent (because they are not separative enough), and the least frequent (because they are too uncommon), and then sampling among the remaining beacons. This set of beacons have translated into $\sim$500 million users. We have observed that the EM algorithm converge in about 25-30 cycles (for hard clustering). This process takes about 15 minutes (for a setting of $\sim$3500 mappers, where the processing takes $\sim$1.5 minutes per mapper; and $\sim$400 reducers, where the processing takes $\sim$4 minutes per reducer) for each cycle in our test cluster, hence a total of 7 hours for the whole process.

~

\section{Results} \label{sec:results}

In this section we will present our evaluation of the proposed clustering algorithm. We will provide both interpretability results (i.e. \emph{how the constructed clusters look like}), as well as numerical results (i.e. \emph{how well advertising with the new clustering algorithm performs}) in the rest of the section. Please note that we are giving a preliminary overview for the performance of our proposed method, and the reasons for non-exhaustive comparisons with other well-known methods are two-folds. First one is the inappropriateness of methods like pLSA and LDA for the dynamic nature of the users in online advertising, as well as the scalability issues in LDA (for the more advanced parallel training methods as mentioned in \S~\ref{sec:previous_work}, we still have significant implementation overhead in terms of man-hours). Second reason is the way we wanted to evaluate the proposed methodology. In our domain, we deploy new models into our actual advertising system to check their performances and any test means lost/gained revenue in such an industrial scenario. In most cases, the online test is the only feasible way to evaluate a new model/method, since this is the only way we can directly measure more relevant metrics (such as revenue, actions/clicks vs. cost of advertising etc.) in online advertising. This is why some of the more recent work~\cite{li_2011} on offline evaluation is often not useful: the metrics that can be examined in such settings (such as click-through rate, AUC etc.) often do not correlate with online metrics. Due to this, it is not often possible to test many models (unless we are fairly confident of the performance) since it may mean lost revenue; this is why we limited the testing of our proposed model and algorithm to our previously employed algorithm based on \emph{cosine similarity} metric. Therefore, in this section, we are basically evaluating how our model replacement efforts resulted.

\begin{figure*}[!t]
\centering
\includegraphics[width=6in]{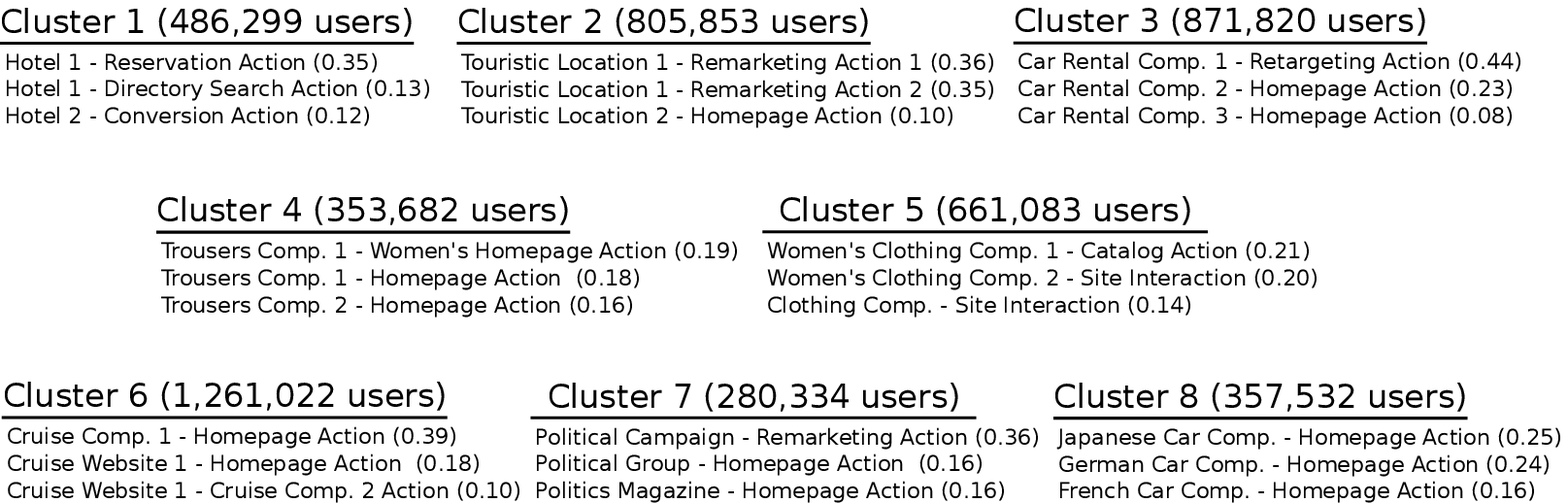}
\caption{Example clusters for the online advertising domain. Cluster ids are followed by the number of users in our system that belong to this cluster. We have listed only the three most significant beacons for each cluster and the numbers in parantheses give the values p(beacon$|$cluster).}
\label{fig_example_clusters}
\end{figure*}

Let us start with our interpretability results, which is a way to manually check whether the clusters generated by our method make sense or not. Figure~\ref{fig_example_clusters} gives a number of example clusters which were constructed by the proposed algorithm. Although we set our system to generate around 600 clusters (due to our previous experience with the domain, we found this to be a good number of partitions for the whole set of users), we only give eight of those clusters for presentational purposes. In the figure, we only show the top three beacons for each cluster, and the number between parantheses after each beacon definition gives the probability of that beacon being generated by a user belonging to that cluster (e.g. p(b$_{\textrm{Hotel 2 - Conversion Action}} ~ | ~ \textrm{Cluster 1}$) $ = 0.12$). Due to privacy reasons, we did not give the exact names of the beacons we used, but rather we gave a categorization on the subject of the beacon and the type of event the beacon indicates. For example, \emph{Homepage Action} means the visit of the home page of an advertiser by a user, whereas a \emph{Conversion Action} usually indicates a purchase. We furthermore give the number of users that had fallen under a specific cluster (during the clustering process) next to the cluster in parantheses (e.g. during the clustering process, 661,083 users out of the $\sim 500$M users we processed fell under \emph{Cluster 5}).

The results, as a subset is given in the figure, shows that the clusters are mostly \emph{meaningful}, i.e. the events (beacons) that fall under the same cluster are similar. Furthermore, it can be seen that they do not always belong to the same advertiser, but usually the same subject. For example, it can be seen that \emph{Cluster 7} have events/beacons from different advertisers under it, one of them being a political campaign, one a political group, and one a political magazine. The clustering algorithm was able to gather these events under the same cluster since they all belong to the subject \emph{Politics} (of course this is the expected behavior since the users that fall under this cluster are the ones that are interested in politics). Similar arguments can easily be made for the other clusters presented in the figure, and most of the clusters that are currently being used in our system, but not presented here.

Next, we will give a comparison of the clustering algorithm against a previous system that was being used at our company's advertising framework again for user segmentation. Our previous algorithm also utilizes user beacons, but employs a \emph{cosine similarity} metric (by taking the beacon set as a feature vector for each user) to determine which users are \emph{closer, or more similar,} to each other, hence should be in the same cluster for action rate and click rate calculation. In this algorithm, each beacon b$_i$ has a weight w$_i$, i.e. $w_i = \alpha |u(b_i)|$ where $u(b_i)$ is the set of users that have fired this beacon at least once. Furthermore, each user $u_j$ is represented by a vector where the entries for the beacons that have been seen by this user has the value 1, and others 0 (i.e. $v(u_j) = [v_1,...,v_n]$ where n is the total number of distinct beacons in the system, and $v_i = 1$ if $u_j$ has seen beacon $b_i$, $0$ otherwise). The algorithm starts with $n$ centroids, where $n$ is the total number of beacons available (furthermore at the beginning, each centroid $c_m$ is a vector $v(c_m)=[v_1, ... , v_n]$ where only $v_m$, representing beacon $b_m$, is 1, and all other entries 0), and applies a k-means on these set of centroids (we take k = 600 similar to the new methodology). Each user $u_j$ is assigned to a centroid, hence cluster, as follows:
\[
Cluster(u_j) = argmax_{c_i \in C} \frac{v(c_i) \cdot v(u_j)}{\|v(c_i)\| ~ \|v(u_j)\|} 
\]
where the norm and dot product calculations take into account the weights ($w_i$) of beacons ($b_i$). Furthermore, the entries for the centroid vectors are updated during the k-means according to vectors of the users assigned to them. In this algorithm, each k-means is followed by a \emph{merging stage} where significantly closer centroids are merged. This process goes until no centroid merges are possible. The disadvantage of this algorithm is two-folds. The first one is the already high (and increasing) number of users in the advertising domain, hence the computational issues. Second one is the assumption that any beacon seen by a user has the same weight for that user (i.e. we do not utilize $p(beacon|user)$ values), which does not take into account the possibility that many of the same events (beacons) may indeed indicate higher interest on a specific type of product.

We have run an A/B test using two models, one utilizing the new clusters/methodology and one the old clusters/methodology. These models have been run on the whole set of campaigns within Turn, where impression traffic was directed to two models with equal priority. In the models, the users that belong to the same cluster have the same predicted action/click rates (which are utilized for calculating the bid value) for the same campaigns. These rates are again calculated from historical data of action/clicks (separately for each campaign) among the users that belong to the same cluster. Therefore, the cluster id is used as a single feature, alongside with the campaign id. We are providing results in terms of \emph{effective cost per action} (eCPA) and \emph{effective cost per click} (eCPC) metrics. These metrics can be described as follows:
\begin{itemize}
\item \textbf{Effective Cost per Action (eCPA):} What is the average amount of money that is spent by an advertiser (on advertising) to receive one action (i.e. purchase etc.)? This metric can be calculated as:
\[
eCPA = \frac{\textrm{Advertising Cost}}{\# \textrm{ of Actions}} ~ .
\]
\item \textbf{Effective Cost per Click (eCPC):} What is the average amount of money that is spent by an advertiser (on advertising) to receive one click (on its ad)? This metric can be calculated as:
\[
eCPC = \frac{\textrm{Advertising Cost}}{\# \textrm{ of Clicks}} ~ .
\]
\end{itemize}
\begin{figure}[htb]
\centering
\includegraphics[width=3.2in]{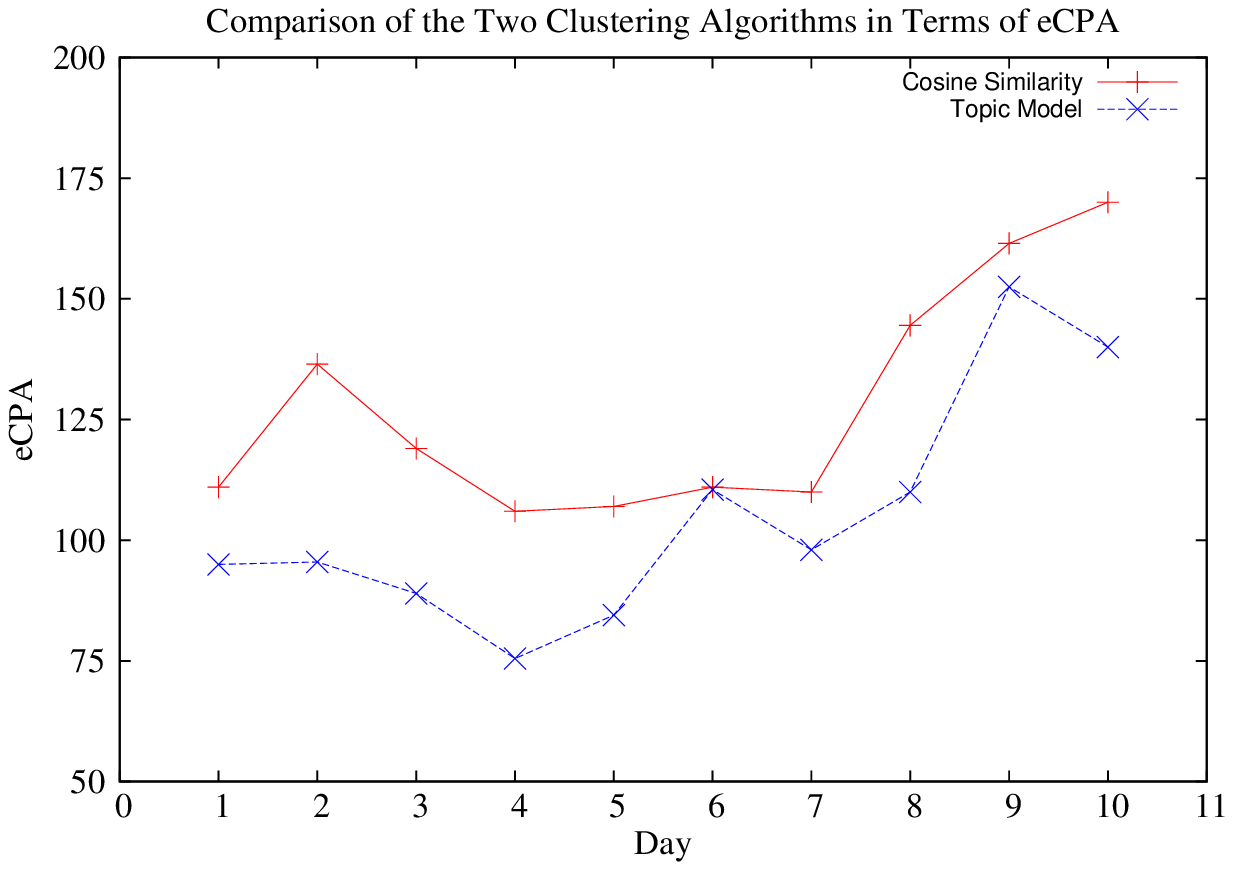}
\caption{Comparison of eCPA Performance for the Clustering Algorithms over 10 Days. Lower eCPA that has been achieved by the proposed methodology indicates better performance.}
\label{fig_eCPA_comp}
\end{figure}
Above metrics are representative of the quality of clustering since our bidding logic takes the cluster ids of users directly into consideration. If we are able to separate the users into meaningful segments (via clustering), then the action probability calculated for each segment is more accurate (therefore our bid values are closer to the actual value of each impression). Due to this, the money spent for each impression (due to bidding) brings more actions/clicks, hence improving eCPC and eCPA metrics.

\begin{figure}[htb]
\centering
\includegraphics[width=3.2in]{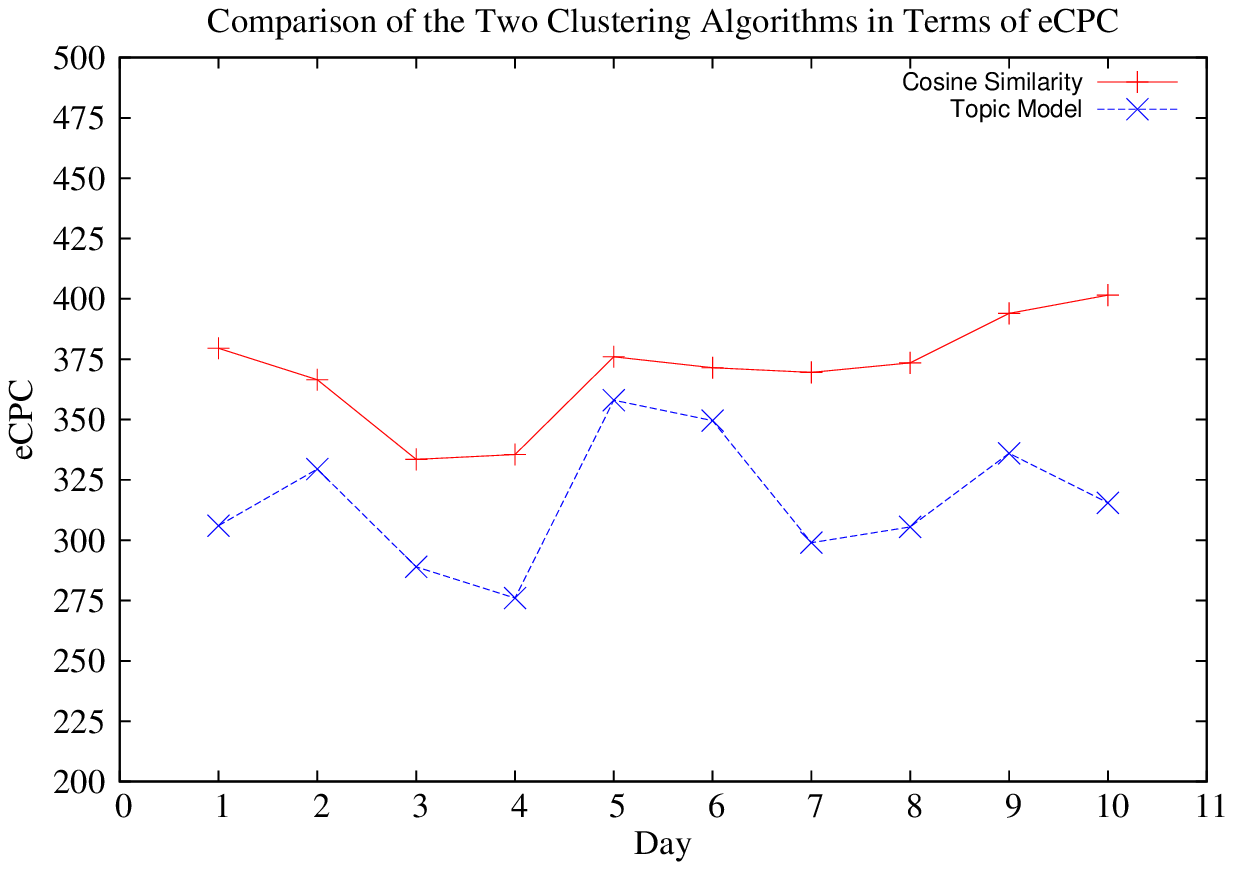}
\caption{Comparison of eCPC Performance for the Clustering Algorithms over 10 Days. Lower eCPC that has been achieved by the proposed methodology indicates better performance.}
\label{fig_eCPC_comp}
\end{figure}

The results of our initial experiments, that span over a 10 day period within July-August 2013, are given in Figures \ref{fig_eCPA_comp} and \ref{fig_eCPC_comp}. While Figure~\ref{fig_eCPA_comp} compares the day-by-day eCPA performance for both algorithms, Figure~\ref{fig_eCPC_comp} presents the same analysis for eCPC performance. Due to privacy issues, we are not presenting the exact values in the plots, but rather we have changed the values by a factor. It can be seen that the proposed algorithm utilizing the topic model performs much better (i.e. lower eCPA or eCPC is better) compared to the algorithm which utilizes the cosine similarity metric. Furthermore, we see that the eCPA has an trend for increasing for both models at the end of the experimentation period. This is due to the \emph{action attribution} problem inherent in advertising systems, where an action that happens is attributed to an ad that is shown to a user several days ago. Due to this, the later days still have some actions that are to be received in time.

\section{Conclusions and Future Work}	\label{sec:conclusion}
In this paper, we described a methodology to cluster users in online advertising. We have explained the difficulties such as non-stable set of users as well as the large data that we have to deal with, and how those factors shape the algorithm we employ or how we implement this algorithm. We have given a brief overview of the implementation in Apache Pig, on Hadoop, as well as some initial experimental results. It is important to mention that we are not claiming to have explored all possible algorithms/models for analysis in this work, but rather that we have developed a meaningful and efficient system that solves a real-world online advertising problem and improves the performance. In summary, we believe that this work fills a significant void in the literature, since it deals directly with the large-scale problems that arise in online advertising domain. Our future work includes the extension of our results, as well as an improved clustering algorithm where the user properties (such as age, gender, location etc.) are used alongside with the item sets (i.e. beacon history) for better determination of clusters.

\bibliographystyle{abbrv}
\bibliography{Geyik_us_cl_arxiv}

\end{document}